\documentclass{article}

\usepackage[numbers]{natbib}

\usepackage{arxiv}

\usepackage{cite}
\usepackage{floatrow}
\usepackage{amsmath,amssymb,amsfonts}
\usepackage{algorithm}
\usepackage{algorithmic}
\usepackage{textcomp}
\usepackage{float}
\usepackage{booktabs}
\usepackage{enumitem}
\usepackage{setspace}
\usepackage[]{graphicx}
\usepackage{amsmath}
\usepackage{amssymb}
\usepackage{epsfig}
\usepackage{epstopdf}
\usepackage{subfigure}
\usepackage{comment}
\usepackage{tabularx}
\usepackage{threeparttable}
\usepackage{adjustbox}
\usepackage{soul}
\usepackage{graphicx}

\usepackage{graphicx}
\usepackage[dvipsnames]{xcolor}

\usepackage{microtype} % Enhances font embedding
\usepackage{cmap}      % Improves PDF searchability and accessibility
\title{\LARGE \bf
Transfer Learning of Tabular Data by Finetuning Large Language Models}

\author{Shourav B. Rabbani,~
        Ibna Kowsar, ~%~\IEEEmembership{}
        Manar D. Samad\\
Department of Computer Science\\
Tennessee State University\\
Nashville, TN, USA\\
\texttt{msamad@tnstate.edu} \\
}

\begin{document}

% \title {Transfer Learning of Tabular Data by Finetuning Large Language Models
 % Contrastive Domain Adaptation by Minimizing Divergence in Source-Target Image Distributions %for out-of-distribution generalization}

\maketitle
%\thispagestyle{empty}
%\pagestyle{empty}

%%%%%%%%%%%%%%%%%%%%%%%%%%%%%%%%%%%%%%%%%%%%%%%%%%%%%%%%%%%%%%%%%%%%%%%%%%%%%%%%
\begin{abstract}

Despite the artificial intelligence (AI) revolution, deep learning has yet to achieve much success with tabular data due to heterogeneous feature space and limited sample sizes without viable transfer learning. The new era of generative AI, powered by large language models (LLM), brings unprecedented learning opportunities to diverse data and domains. This paper investigates the effectiveness of an LLM application programming interface (API) and transfer learning of LLM in tabular data classification. LLM APIs respond to input text prompts with tokenized data and instructions, whereas transfer learning finetunes an LLM for a target classification task. This paper proposes an end-to-end finetuning of LLM to demonstrate cross-data transfer learning on ten benchmark data sets when large pre-trained tabular data models do not exist to facilitate transfer learning. The proposed LLM finetuning method outperforms state-of-the-art machine and deep learning methods on tabular data with less than ten features - a standard feature size for tabular data sets. The transfer learning approach uses a fraction of the computational cost of other deep learning or API-based solutions while ensuring competitive or superior classification performance.

\end{abstract}

%%%%%%%%%%%%%%%%%%%%%%%%%%%%%%%%%%%%%%%%%%%%%%%%%%%%%%%%%%%%%%%%%%%%%%%%%%%%%%%%

\section{Introduction}

In the past decade, the field of artificial intelligence (AI) has advanced to new eras with the advent of deep learning (DL) and large language models (LLMs). While DL methods have revolutionized image learning and computer vision applications~\citep{ALAM2020}, pre-trained LLMs have shown unprecedented opportunities for generative AI. Image and text are primary subjects of AI because vision and language are two main modes of human communication. In contrast, tabular data structured in rows and columns are ubiquitous in science, engineering, and industrial applications, which are yet to utilize the strengths of deep learning~\citep{kowsar2023buet, rabbani2024gceals}. Despite unprecedented achievements of deep learning, traditional machine learning remains \emph{de facto} for tabular data. Recent literature has concluded that deep learning methods are yet to “conquer the castle” of tabular data~\citep{Kadra2021} or deep learning is “not all we need” for tabular data~\citep{Shwartz-Ziv2022}. With only a fraction of samples and features of image or text data sets, most tabular data sets are often not considered for data-hungry deep learning.

The limited sample size problem in data-hungry deep learning is investigated by zero-shot or few-shot learning methods~\citep{data_hungry_few_shot1, nam2023stunt}. Recent studies show promising performance of LLMs in zero or few shot learning of tabular data on tasks like classification~\citep{Han2024FeatLLM, Huertas2024ImproveLGMonFewShots, tabllm}, reasoning~\citep{shi2024ehrReasoning, li2024cancergpt}, and text generation~\citep{shi2024ehrReasoning, slack2023tablet}. These recent studies show that a pre-trained LLM can perform diverse tasks beyond text generation using few or no training samples. One key advantage of an LLM is its ability to learn from text semantics. While deep and machine learning techniques are designed to learn from numerical or encoded categorical features, LLMs can additionally learn semantics from feature names and descriptions in data sets or human instructions. Emerging studies on tabular data show that the efficacy of LLM-based application programmable interface (API) is limited to zero-shot or few-shot learning~\citep{Han2024FeatLLM}. APIs are ready to take text-based human instructions to generate results without any learning steps. However, it is unknown how LLM API's performance compares against the finetuning of a pre-trained language model in tabular data classification. We hypothesize that an end-to-end finetuning of a generative language model will overcome the data processing and performance limitations of popular LLM APIs and deep learning methods. The classification performance of LLMs is compared against baseline deep and machine learning methods on ten benchmark tabular data sets.

The organization of the paper is as follows. Section \ref{sec:background} provides a background on the challenges of deep learning of tabular data and LLM approaches proposed for learning tabular data, and limitations of API-based generative models. Section \ref{sec:methods} describes the proposed computation framework developed to finetune a generative language model to classify tabular data. Section \ref{sec:results} shares the key findings with some discussions of the results. The paper concludes in Section \ref{sec:conclusions}.

\begin{figure*}[t]
\centering
\includegraphics[width=1\textwidth]{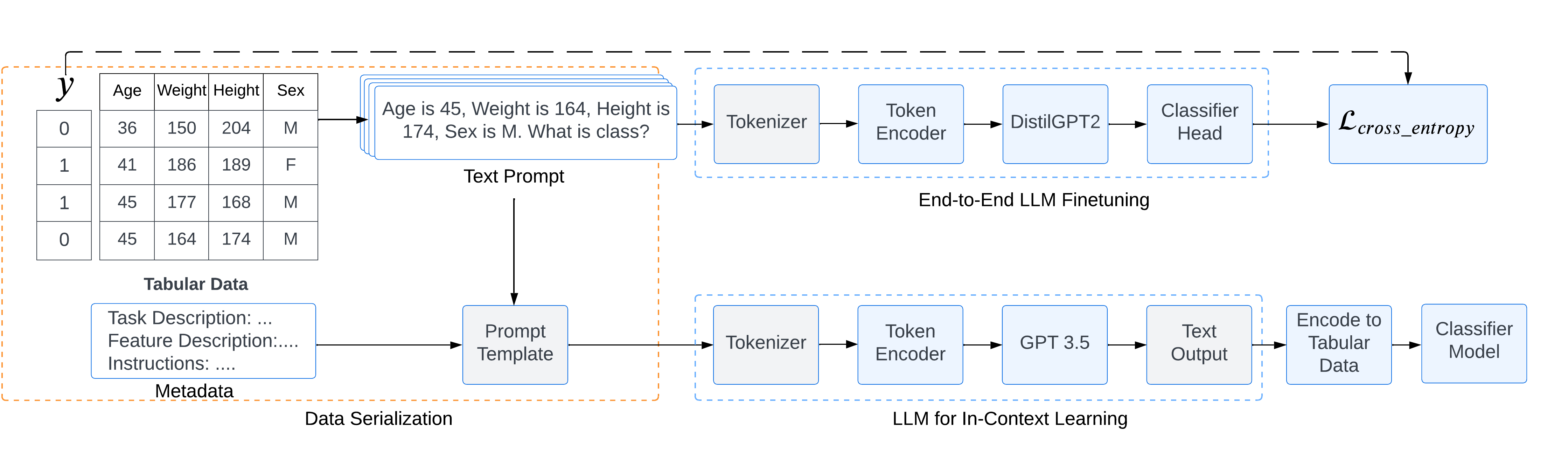}
\vspace{-20pt}
\caption{
Tabular data transfer learning framework using a large language model (LLM).  In-context learning using LLM includes a detailed text prompt and is without a classifier head or finetuning step.
}
\label{fig:llm-pipline}
\end{figure*}

\section{Background}
\label{sec:background}

Data structured in tables, namely tabular data, appear as spreadsheets or petabytes of data stored in relational databases in science, business, industry, and medicine. The feature space of tabular data is heterogeneous (e.g., salary, age) and permutation invariant, which fundamentally differs from homogeneous (e.g., pixel) image or text data where word or pixel positions are significant. The effectiveness of deep learning methods in AI can be attributed to homogeneous features sequentially positioned in pixel space or sentences. In contrast, recent studies reveal that deep learning methods are often less effective than traditional machine learning on tabular data with heterogeneous feature space~\citep{Kohler2019, Shwartz-Ziv2022, Borisov2021}. 

A large sample size is an important requirement of deep learning. The 100 most downloaded benchmark tabular data sets from the popular University of California Irvine (UCI) machine learning data repository have a median sample size of 660 and feature size of 18~\citep{Abrar2023icaci}. Therefore, deep learning studies on tabular data selectively choose data sets with large sample sizes ($>$10,000)~\citep{Borisov2021, Gorishniy2021ftt, yan2023_t2g}. Alternatively, zero-shot and few-shot deep learning methods have been proposed to demonstrate the economical use of limited data samples when a large pre-trained model is available. An LLM has enabled no or few shot learning across various data domains, including tabular data.

The LLM literature on tabular data can be broadly grouped as in-context learning and fine-tuning~\citep{Fang2024LLMSurvey}. In-context learning uses prompt engineering by sending custom prompts to guide the LLM in generating the desired response~\citep{brown2020languagemodelsfewshotlearners}. For example, the prompt for tabular classification includes data samples inscribed in text with tasks and additional descriptions~\citep{Han2024FeatLLM, tabllm}. 
Prompt engineering aims at effective user interaction with the API at a higher abstraction level, avoiding the need to access or modify the underlying model. For example, the FeatLLM method prompts an LLM API to create rules or conditions on important features to facilitate the tabular data classification task. Unfortunately, the API has an upper limit on the number of tokens in the text prompt. For example, the gpt-3.5-turbo-0613 model from OpenAI can accept text prompts with a maximum of 16,385 tokens, which limits the number of input samples. Therefore, the API can process a fraction of available samples, which may not fully utilize data with a larger sample size. In contrast, the fine-tuning approach learns from the existing knowledge of an LLM by appending a classifier head to perform a custom classification task~\citep{tabllm, Dinh2022Lift}. The model weights can be finetuned via backpropagation following the loss function specific to a learning task. 

The current literature is unclear on how the in-context solutions via an API compare against the LLM finetuning approaches in tabular data classification. The limit on token sizes has not shown the full capacity of LLM beyond no and few-shot learning. 
This paper proposes an end-to-end finetuning of an LLM to perform transfer learning of tabular data. The classification performance and computational costs are compared against traditional machine and deep learning methods. The proposed method demonstrates a transfer learning solution for tabular data, especially when no large pretrained tabular data models exist to facilitate transfer learning.

\section {Methods}
\label{sec:methods}
This section provides the computational steps in preparing tabular data for classification using LLMs in addition to the baseline methods and experimental steps.

\subsection{LLM for tabular data classification}

A tabular data set is structured as $n \times m$ dimensional matrix where $n$ and $m$ are sample and feature dimensions, respectively. LLMs can not directly process tabular data structured in rows and columns. Therefore, each row sample of a table must be converted to a text prompt. The steps to perform tabular data classification using an LLM are provided below.

\subsubsection{Data serialization}
 
Data serialization uses a computer program to textualize each sample with feature names and values in a text prompt.  For example, a data sample with age and sex features is converted to ``Age is 25. Sex is male." Additional descriptions of features and tasks can be included in the text prompt as metadata, which is known as prompt engineering. Metadata provides important task-specific contexts to enhance the in-context learning of a pre-trained LLM.  We perform data serialization in two steps. First, the feature and value pairs of each sample are separated into $n$ serialized tokens. Second, additional information in the text prompt, including feature-specific metadata and a task, is included, as shown in Figure~\ref{fig:llm-pipline}.

\subsubsection{Data tokenization}
After data serialization, text samples are tokenized by splitting into smaller components (e.g., words, subwords, or characters). Each unique token is a part of the model's vocabulary and mapped to a learnable $d$-dimensional embedding. Input tokens of each sample are padded or truncated to maintain a constant sequence length in line with the LLM's input shape requirements.  In-context learning using an API limits the number of input tokens due to memory constraints. However, DistilGPT2 can take up to 1024 tokens per sample without restrictions on the sample size.

\subsubsection{Large language models}
We use DistilGPT2~\citep{distill_gpt2} for transfer learning, which is a lightweight and open-source large language model obtained from the HuggingFace project~\citep{huggingface_distilgpt2}. DistilGPT2 is a knowledge distillation version of ~GPT-2 with 82 million parameters. The DistilGPT2 model is trained using the English Wikipedia and Toronto Book Corpus ~\citep{zhu2015}. The authors of DistilGPT2 show that it retains 97\% of full GPT-2 performance with 40\% fewer parameters. DistilGPT2 has six transformers designed to take inputs from a token encoder. The token encoder converts input tokens to continuous vector representations in high-dimensional space. For in-context learning, we use OpenAI's gpt-3.5-turbo-0125 model to evaluate its performance in a recently proposed method for tabular data, FeatLLM~\citep{Han2024FeatLLM}.

\subsubsection{Modeling for classification}
An LLM converts a batch of tokenized sequences into a new feature representation. A classifier head transforms the LLM feature representation into a logit vector to show the class correspondence. The LLM with a classifier head is finetuned in one of two ways: 1) LLM with all frozen weights and 2) end-to-end learning involving all trainable LLM weights. The classification task in finetuning minimizes a cross-entropy loss, as shown in Equation~\ref{eq:ce}.
%%%%%%%%%%%%%%%%%
\begin{equation}\label{eq:ce}
\mathcal{L}_{\text{CE}} = - \frac{1}{n} \sum_{i=1}^{n} \sum_{c=1}^{C} y_{ic} \log \hat{p}_{ic}
\end{equation}
Here, $n$ is the number of samples in a batch, $C$ is the number of classes, $y_{ic}$ is the true label for the $i$-th sample and class $c$, and $\hat{p}_{ic}$ is the predicted likelihood of class $c$ for the $i$-th sample. 

In contrast, in-context learning involves no task-specific classifier head or learning steps. For example, FeatLLM prompts the LLM to output decision rules related to the classification task on important features, which are used to create a binary data matrix for standalone classification.

\subsection{Baseline models}
The baseline classification methods include the GBT classifier, an MLP, and a state-of-the-art deep learning method proposed for tabular data, self-supervised contrastive learning using random feature corruption (SCARF)~\citep{bahri2022} implemented in~\citep{rabbani2024AttVsCont}. The GBT model is implemented using the Scikit-learn package~\citep{sklearn}. The MLP network comprises three fully-connected hidden layers with 128, 64, and 32 neurons. We have updated the prompt generator of the FeatLLM method to include as many samples as possible under the constraint on token size. 

\begin{table}[h]
\centering
\caption{Tabular data sets used to evaluate baseline and proposed LLM-based transfer learning methods. UCI: The UCI Machine Learning Repository, OpenML: OpenML Data Repository, and MIMIC: Medical Information Mart for Intensive Care. Shots are specific to the in-context learning using FeatLLM\citep{Han2024FeatLLM}.}
\label{tab:datasets}
\scalebox{0.67}{
\begin{tabular}{llrrrlr}
\toprule
 Data set & Source & Samples & Features & Classes & Difficulty & Shots  \\
\midrule
Dermatology & UCI & 366 & 35 & 6 & easy & 42 \\
Breast-cancer & UCI & 569 & 31 & 2 & easy & 48 \\
Diabetes & UCI & 768 & 9 & 2 & easy & 238 \\
Heart & UCI & 918 & 12 & 2 & easy & 190 \\
Credit-g & UCI & 1000 & 21 & 2 & easy & 105 \\ \midrule
Blood-transfusion & OpenML & 748 & 5 & 2 & hard & 450 \\
Pc3 & OpenML & 1563 & 38 & 2 & hard & 40 \\
Steel-plate-fault & OpenML & 1941 & 28 & 7 & hard & 49 \\
Kc1 & OpenML & 2109 & 22 & 2 & hard & 82 \\
Sepsis & MIMIC & 2164 & 44 & 2 & hard & 36 \\
\bottomrule
\end{tabular}
}
\end{table}

\subsection{Experimental setup}
The experiments are performed on an Ubuntu 20.04 machine with an Intel(R) Core (TM) i9-13900F CPU @ 5.60GHz with 32 cores, 64GB RAM, and an RTX 4090 24GB GPU. Since we utilize PyTorch modules to build and train our model for deep learning methods, it automatically uses multiple CPUs for parallel processing. A five-fold cross-validation scheme is used for all experiments, where four folds are used to train a model and test using the left-out fold. One-eighth of the training data is used to validate the model against overfitting. We follow the experimental setup outlined in~\citep{rabbani2024AttVsCont} while running GPT, MLP, and SCARF methods.

DistilGPT2 is trained for up to 100 epochs with an early-stopping criterion and patience 10 using tokenized inputs. The model training uses a weight decay of 0.01 and a learning rate of 0.00005. The input batch size is 16. The final LLM embedding of size 768 is mapped to class logits for classification. The best classification model on the validation loss is saved for subsequent evaluation on test data folds. We report the average area under the operating characteristic curve (AUC) across five folds to report the final classification performance.

\section {Results}
\label{sec:results}
The findings related to model design and experiments are discussed below.

\subsection{Tabular data sets}
The proposed LLM-based solution, deep and machine learning methods are evaluated using ten tabular data sets of diverse domains and sample sizes, including an electronic health record (EHR) data set, as shown in Table~\ref{tab:datasets}. We select an equal number of hard and easy-to-classify data sets based on a data selection criterion used in ~\citep{rabbani2024AttVsCont}. The ten tabular data sets are obtained from the UCI machine learning~\citep{Dua_2019uciML} and  OpenML~\citep{OpenML2013} repositories. The EHR data set is sourced from the Health Gym project, which is curated from the MIMIC-III database~\citep{kuo2022health}.  The metadata of each data set is used to complete a prompt template with the classification tasks and feature description as required by FeatLLM. The maximum number of samples (shots) for in-context learning of FeatLLM, due to the 15000 limit on token size, is shown in Table~\ref{tab:datasets}.
\begin{figure}[ht]
    \centering
    \includegraphics[width=0.70\textwidth]{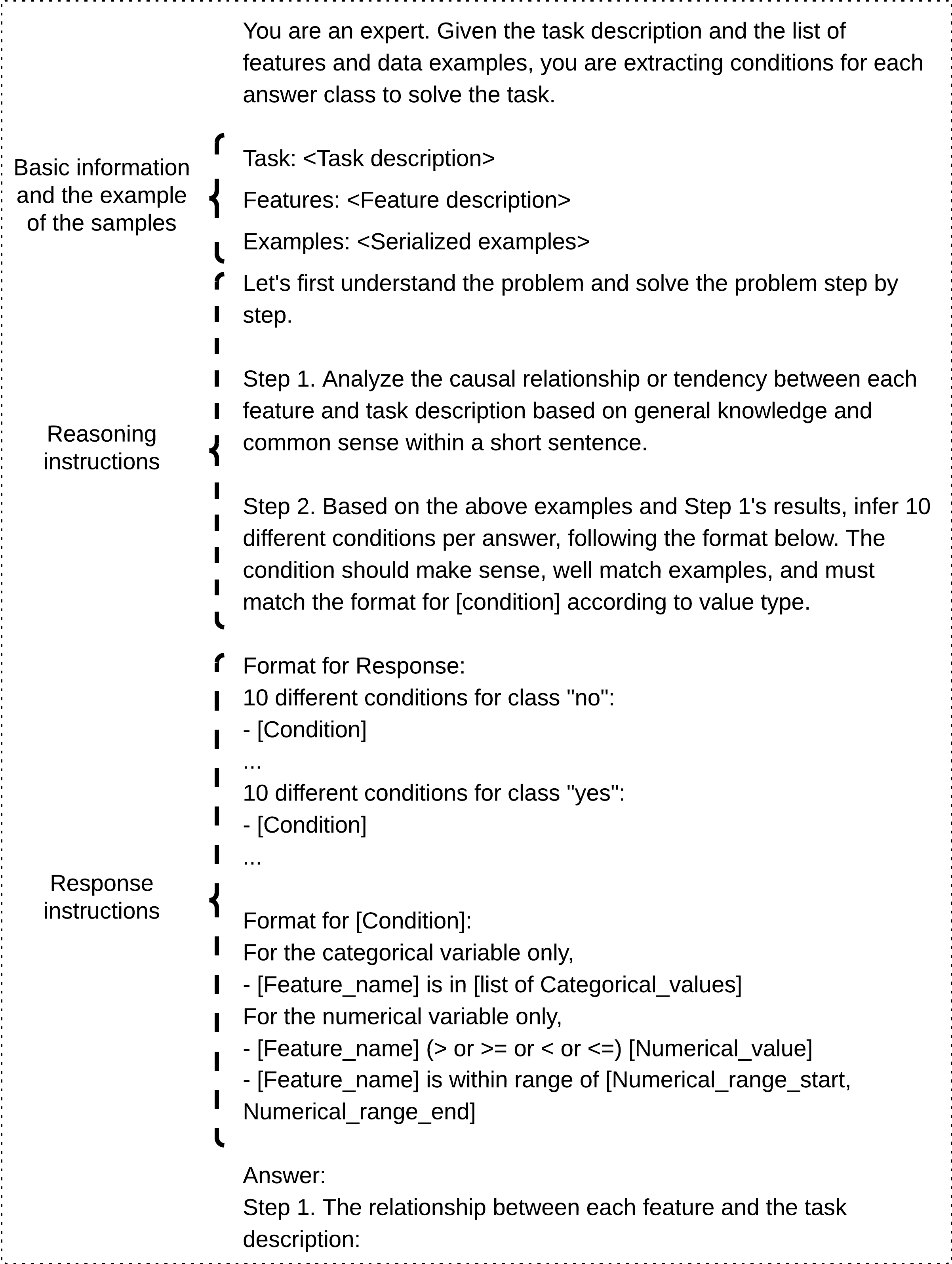}
    \caption{Text prompt template for in-context learning using FeatLLM~\citep{Han2024FeatLLM}.
    }
    \label{fig:featllm-template}
\end{figure}
\begin{figure}[ht]
    \centering
    \includegraphics[width=0.70\textwidth]{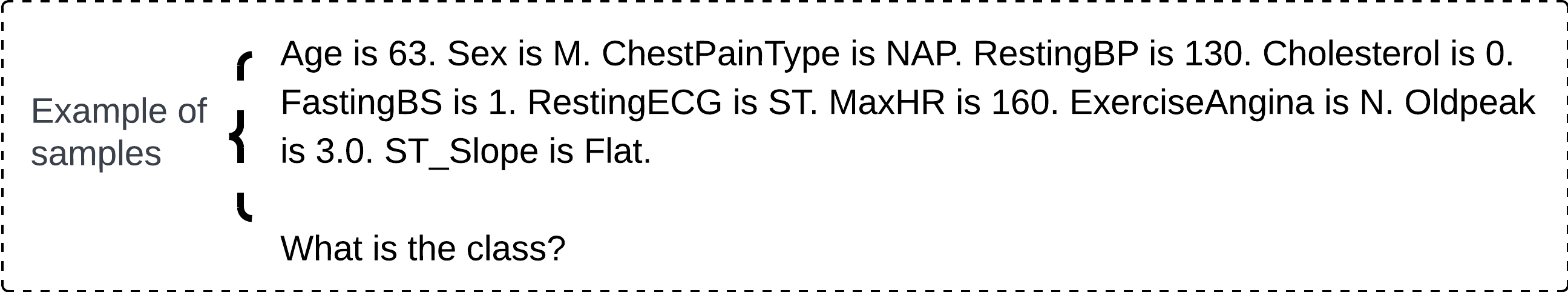}
    \caption{Text prompt template for transfer learning using DistilGPT2~\citep{distill_gpt2}.
    }
    \label{fig:feature-text-template}
\end{figure}
\subsection{Prompt engineering}
In-context learning has a limit on the number of tokens in input prompts. Therefore, the size of FeatLLM's output is limited to 1385 to reserve 15000 tokens for the input prompt. The FeatLLM prompt template also includes reasoning instructions, response instructions, feature metadata, and serialized examples, as shown in Figure~\ref{fig:featllm-template}. In contrast, transfer learning of LLM has a 1024-token limit per sample and can process data in batches to train on the entire data set. The transfer learning method uses a straightforward feature-to-text serialization method, avoiding the need for a complex text prompt, as presented in Figure~\ref{fig:feature-text-template}.
\begin{figure}[t]
\vspace{-10pt}
    \centering
    \subfigure[All LLM weights are trainable]{
    \includegraphics[width=0.6\textwidth]{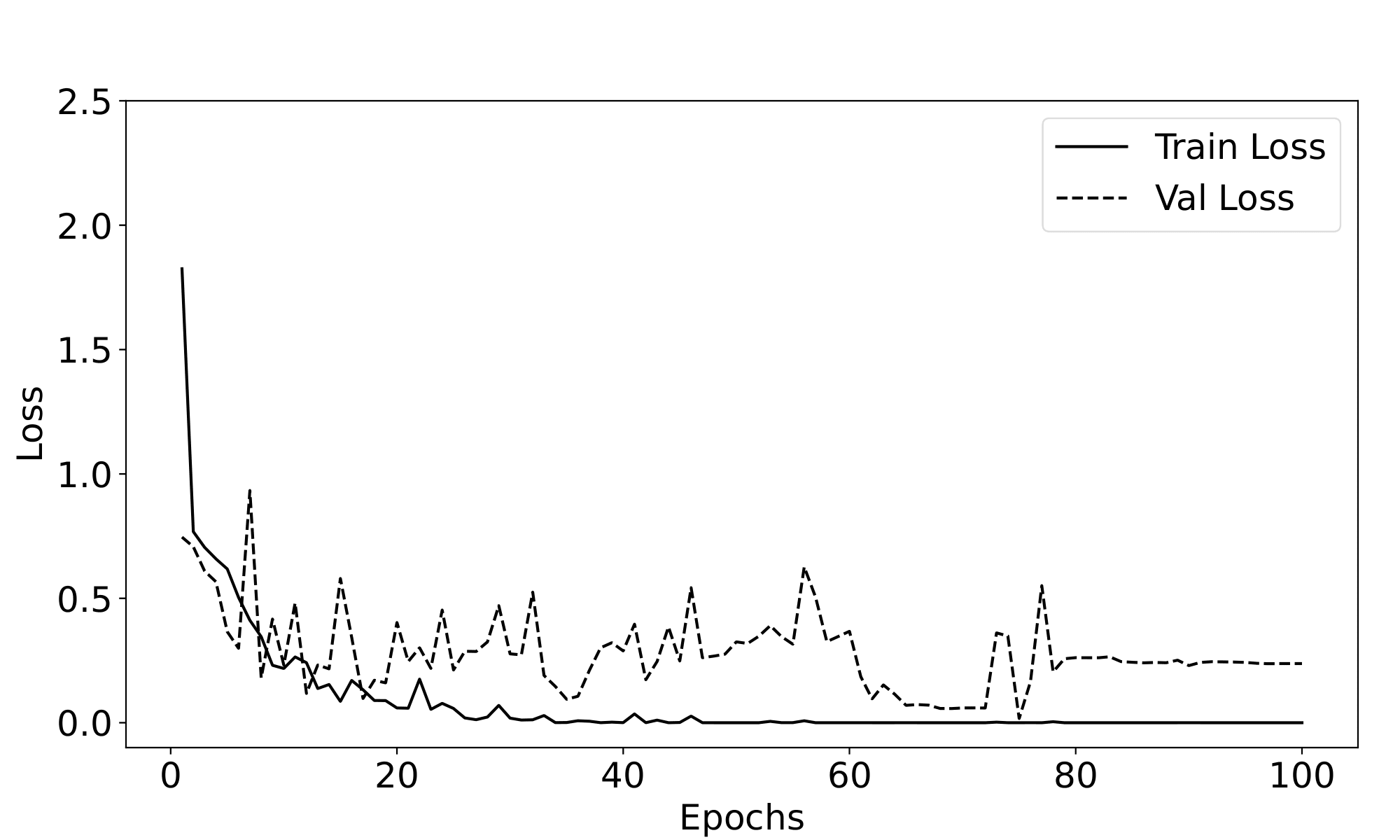}}
    % \hfill
    \setlength{\subfigtopskip}{0.001cm}
  %  \hfill
   % \vspace{-20pt}
    \subfigure[All LLM weights are frozen]{
    \includegraphics[width=0.6\textwidth]{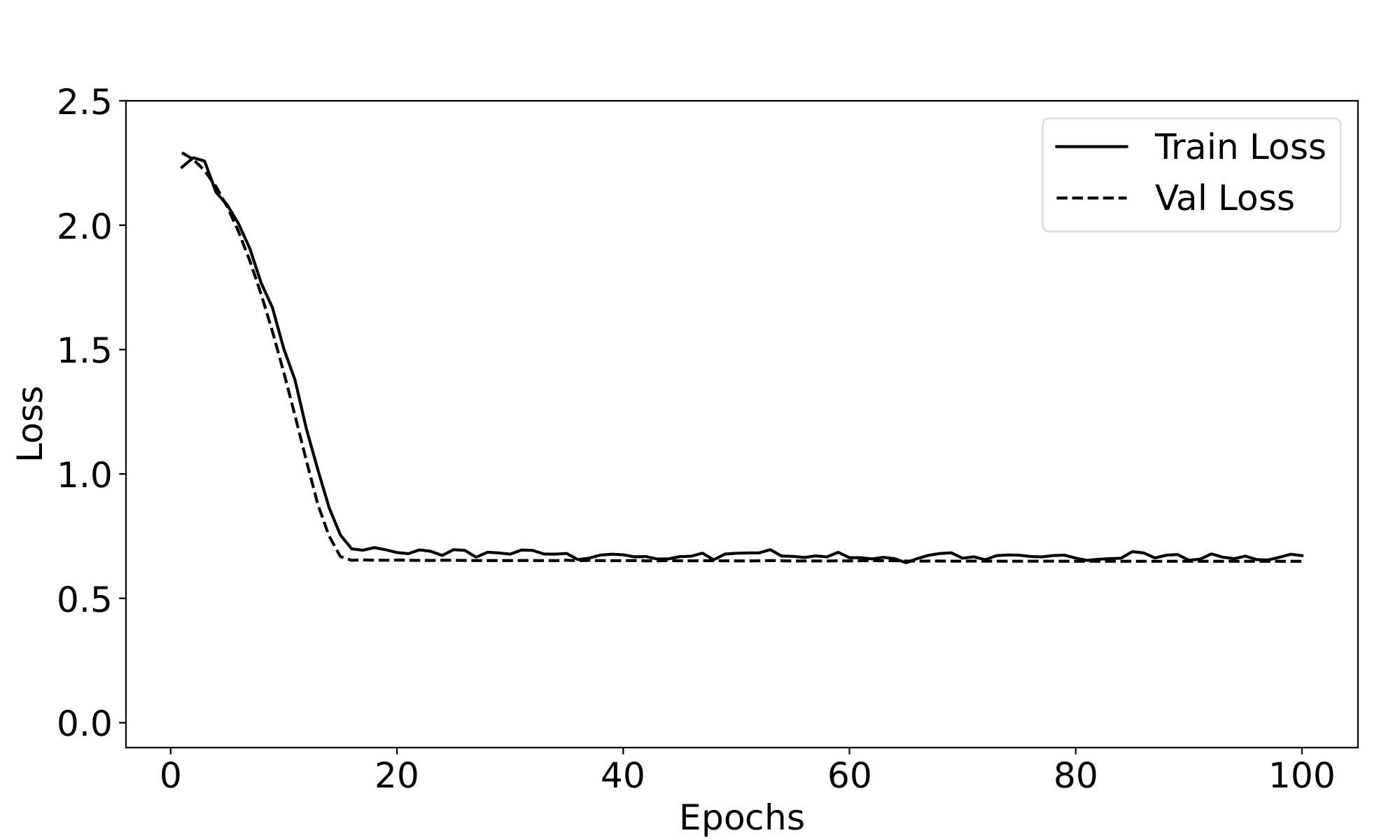}}
    \vspace{-10pt}
    \caption{ Convergence plots of proposed transfer learning of tabular data using a large language model (LLM)}% \sbr{[SBR: The images themselves have padding. Either tighten them up via code or use trim parameters in latex]}}
    \label{fig:loss_curves}
\end{figure}
\subsection{Model training} In transfer learning, an LLM with all frozen weights is more rigid than the model with all trainable weights. The loss curves of the LLM with all frozen weighs reach a plateau above the loss value of 0.5, indicating underfitting. In contrast, LLM with all trainable weights yields a training loss closer to zero but starts to overfit after 30 epochs, as shown in Figure~\ref{fig:loss_curves}. Our early stopping criterion selects the model with the lowest validation loss for subsequent testing. Therefore, the proposed LLM-based transfer learning needs a proper balance of frozen and trainable weights to optimize the classification performance.

\subsection {Model performance}
\begin{table*}[!t]
    \centering
\scalebox{0.8}{
\begin{tabular}{lcccccc}
\toprule
Data set & GBT & MLP & SCARF~\citep{bahri2022} & FeatLLM~\citep{Han2024FeatLLM} & \shortstack{Frozen\\Weights}  & \shortstack{Finetuning\\End-to-end}\\

\midrule
Dermatology & 0.999 (0.001) & 0.999 (0.001) & 0.999 (0.002) & 0.983 (0.009) & 0.568 (0.041) & 0.992 (0.010) \\
Breast-cancer & 0.992 (0.006) & 0.992 (0.008) & 0.994 (0.008) & 0.987 (0.009) & 0.654 (0.097) & 0.976 (0.014) \\
Diabetes & 0.828 (0.024) & 0.792 (0.029) & 0.755 (0.025) & 0.829 (0.033) & 0.637 (0.082) & 0.820 (0.032) \\
Heart & 0.922 (0.022) & 0.909 (0.031) & 0.893 (0.017) & 0.908 (0.021) & 0.872 (0.043) & 0.917 (0.025) \\
Credit-g & 0.776 (0.025) & 0.761 (0.042) & 0.747 (0.036) & 0.670 (0.037) & 0.571 (0.077) & 0.755 (0.030) \\
Blood-transfusion & 0.711 (0.039) & 0.762 (0.032) & 0.701 (0.031) & 0.733 (0.021) & 0.661 (0.069) & 0.738 (0.026) \\
Pc3 & 0.837 (0.028) & 0.776 (0.056) & 0.783 (0.022) & 0.751 (0.025) & 0.789 (0.025) & 0.798 (0.016) \\
Steel-plate-fault & 0.964 (0.007) & 0.922 (0.008) & 0.939 (0.009) & 0.866 (0.014) & 0.736 (0.008) & 0.897 (0.009) \\
Kc1 & 0.814 (0.021) & 0.768 (0.032) & 0.688 (0.036) & 0.786 (0.024) & 0.791 (0.018) & 0.790 (0.022) \\
Sepsis & 0.959 (0.005) & 0.972 (0.005) & 0.976 (0.007) & 0.655 (0.032) & 0.618 (0.070) & 0.873 (0.019) \\
\midrule
Easy data sets & 2.0 (0.8) & 2.8 (1.0) & 3.0 (2.3) & 3.5 (1.7) & 6.0 (0.0) & 3.5 (1.3) \\
Hard data sets & 1.8 (1.3) & 3.0 (1.7) & 3.7 (1.9) & 4.7 (1.0) & 4.8 (1.8) & 3.0 (0.9) \\ \midrule 
Avg. Rank & 1.9 (1.1) & 2.9 (1.4) & 3.4 (2.0) & 4.2 (1.4) & 5.3 (1.5) & 3.2 (1.0) \\
\bottomrule
\end{tabular}
}
    \caption{Average area under the receiver operating characteristic curves (AUC) across five cross-validation folds. Frozen denotes when all LLM parameters are frozen during finetuning. End-to-end denotes when all parameters are trainable. Average classification ranks are shown for easy and hard data set categories. }
    \label{tab:results}
\end{table*}
Table~\ref{tab:results} presents the classification performances of the proposed LLM-based transfer learning along with machine and deep learning methods. In terms of rank order, the traditional machine learning method (e.g., GBT) appears to be the best performing method, which is also supported by the tabular data classification literature.  

Although in-context learning via FeatLLM uses one of the most up-to-date LLMs, its performance may be significantly limited by the maximum token size. In-context learning via LLM API does not involve any finetuning or transfer learning. Without transfer learning, text input prompts (0.733 (0.021)) are able to achieve a better performance than the best overall GBT model (0.711(0.039)) on the blood transfusion data set. The shot and sample ratio (ssr) is 60\% for this data set. The performance of in-context learning is on par with GBT on the dermatology (ssr: 11.5\%), breast cancer (ssr: 8.4\%), and diabetes (31\%) data sets. The LLM may have some prior knowledge about these medical domains or data sets, which may have contributed to promising results despite few-shot learning.

Transfer learning using an LLM with all frozen weights (Frozen) appears as the worst classification solution for tabular data. The performance drop may be attributed to the distilled LLM compared to the one used for in-context learning with a detailed input prompt. The rigidness of frozen weight is then fully relaxed to perform end-to-end finetuning of the LLM (End-to-end), which has substantially improved the average rank ordering from 6.0 (0.0) to 3.5 (1.3) on easy data sets and from 4.8 (1.8) to 3.0 (0.9) on hard data sets. Overall, end-to-end finetuning of all trainable LLM weights yields the best rank ordering among the LLM approaches. Similar to FeatLLM, this transfer learning approach outperforms the overall best GBT method on the blood transfusion data set. The blood transfusion data set has the lowest number of features (five) among all data sets. The diabetes data set has the second lowest number of features (nine). LLM solutions show diabetes classification performance on par or superior to the overall best classification method, GBT. Therefore, transfer learning via LLM can be the best solution when the feature size is less than ten, which is very common in tabular data sets.

End-to-end transfer learning of an LLM outperforms the state-of-the-art deep learning model for tabular data (SCARF) on six of ten data sets with superior average rank ordering. In addition to competitive classification performance, LLM-based finetuning demonstrates the feasibility of cross-data (text-to-tabular) transfer learning, where a pretrained large text model can learn tabular data for classification. It is noteworthy that transfer learning using large pretrained models is one of the cornerstones of modern AI. While large pretrained models for text and images are widely available, no such models exist for tabular data. Therefore, the proposed transfer learning solution will inspire future research to reap the full benefits of deep learning in learning tabular data.

\subsection{Computational costs}
Table \ref{tab:time_taken} presents the average time in seconds to complete four-fold training and one-fold testing. As expected, GBT and MLP are the most efficient methods for completing these tasks within ten seconds. The in-context learning method (FeatLLM) suffers from long execution times of up to five minutes due to network latency from API access. Transfer learning using LLM, either with frozen or fully trainable model weights, takes 10\% to 50\% of the computational cost required in in-context learning or deep learning methods for tabular data. Considering the trade-off between computational costs and classification accuracy, transfer learning of tabular data using LLM stands out as the best learning approach for tabular data.

\begin{table}[]
\centering
\caption{Average computational time in seconds to complete four-fold training and one-fold testing.}
\label{tab:time_taken}
\scalebox{0.9}{
\begin{tabular}{lrrrr}
\toprule
Time (s) & Heart & Credit-g & Steel-plate-fault & Sepsis \\
\midrule
GBT & 0.1 & 0.2 & 7.6 & 3.0 \\
MLP & 5.0 & 5.0 & 7.7 & 9.6 \\
SCARF & 539.8 & 552.8 & 780.7 & 885.3 \\
FeatLLM & 287.4 & 382.0 & 761.9 & 321.5 \\
Frozen & 53.8 & 99.2 & 361.2 & 449.4 \\
End-to-end & 52.3 & 75.2 & 267.3 & 419.1 \\
\bottomrule
\end{tabular}
}
\end{table}

\section{Conclusions}
\label{sec:conclusions}
This paper investigates the performance of in-context learning and transfer learning of LLM in tabular data classification tasks. Our results reveal that transfer learning using LLM is superior to in-context LLM learning and deep learning methods proposed for tabular data in terms of performance and computational costs. This paper shows one of the unique strengths of LLMs in text-to-tabular transfer learning, which may advance computing in data domains with limited samples or where large pre-trained models like image and text are not feasible. In the future, LLMs tailored to the learning requirements and knowledge of tabular data may advance the field of data science.

\section {Acknowledgement}

 The research reported in this publication was partially supported by the US National Science Foundation (NSF) award \# 2431058 and received support from the Air Force Office of Scientific Research under Grant Number W911NF-23-1-0170. The content is solely the responsibility of the authors and should not be interpreted as representing the official policies, either expressed or implied, of the Army Research Office or the U.S. Government.

\bibliographystyle{unsrt}
\bibliography{myBib}

\end{document}